\documentclass[letterpaper, onecolumn, 10 pt, conference]{ieeeconf}  

\IEEEoverridecommandlockouts                              

\usepackage{amsmath,amsthm,amssymb}
\usepackage{graphicx,fixltx2e}
\usepackage{hyperref}
\usepackage{xy}
\usepackage[]{algorithm2e}
\usepackage{bbm}
\usepackage{todonotes}
\usepackage{float}
\usepackage{cite}

\newcommand{\R}{\mathbb{R}}

\newcounter{subeqn} %

\newcommand{\Exp}{\mathbb{E}}

\def\BibTeX{{\rm B\kern-.05em{\sc i\kern-.025em b}\kern-.08em
    T\kern-.1667em\lower.7ex\hbox{E}\kern-.125emX}}

\makeatletter
\newcommand{\pushright}[1]{\ifmeasuring@#1\else\omit\hfill$\displaystyle#1$\fi\ignorespaces}
\newcommand{\pushleft}[1]{\ifmeasuring@#1\else\omit$\displaystyle#1$\hfill\fi\ignorespaces}
\makeatother

\makeatletter
\def\@citex[#1]#2{\leavevmode
\let\@citea\@empty
\@cite{\@for\@citeb:=#2\do
{\@citea\def\@citea{,\penalty\@m\ }%
\edef\@citeb{\expandafter\@firstofone\@citeb\@empty}%
\if@filesw\immediate\write\@auxout{\string\citation{\@citeb}}\fi
\@ifundefined{b@\@citeb}{\hbox{\reset@font\bfseries ?}%
\G@refundefinedtrue
\@latex@warning
{Citation `\@citeb' on page \thepage \space undefined}}%
{\@cite@ofmt{\csname b@\@citeb\endcsname}}}}{#1}}
\makeatother

\allowdisplaybreaks

\begin{document}



\title{\LARGE \bf
Real-Time Risk-Bounded Tube-Based Trajectory Safety Verification
}

\newcommand{\CH}[1]{\textcolor{red}{[CH: #1]}}
\newcommand{\AW}[1]{\textcolor{red}{[AW: #1]}}
\newcommand{\AJ}[1]{\textcolor{red}{[AJ: #1]}}
\let\oldemptyset\emptyset
\let\emptyset\varnothing
\author{Ashkan Jasour*, Weiqiao Han*%
, and Brian Williams
\thanks{All authors are with the Computer Science and Artificial Intelligence Laboratory (CSAIL), Massachusetts Institute of Technology (MIT). \{jasour,weiqiaoh,williams\} @ mit.edu.
This work was partially supported by the Boeing grant 6943358. *These
	authors contributed equally to the paper.} 
}

\maketitle

\begin{abstract}
In this paper, we address the real-time risk-bounded safety verification problem of continuous-time state trajectories of autonomous systems in the presence of uncertain time-varying nonlinear safety constraints. Risk is defined as the probability of not satisfying the uncertain safety constraints. Existing approaches to address the safety verification problems under uncertainties either are limited to particular classes of uncertainties and safety constraints, e.g., Gaussian uncertainties and linear constraints, or rely on sampling based methods. 
In  this  paper,  we  provide  a fast  convex algorithm  to  efficiently  evaluate  the  probabilistic nonlinear safety constraints  in  the  presence  of arbitrary probability distributions and long planning horizons in real-time, without the need for uncertainty samples and time discretization. The provided approach verifies the safety of the given state trajectory and its neighborhood (tube) to account for the execution uncertainties and risk. In the provided approach, we first use the moments of the probability distributions of the uncertainties to transform the probabilistic safety constraints into a set of deterministic safety constraints. We then use convex methods based on sum-of-squares polynomials to verify the obtained deterministic safety constraints over the entire planning time horizon without time discretization. To illustrate the performance of the proposed method, we apply the provided method to the safety verification problem of self-driving vehicles and autonomous aerial vehicles.

\end{abstract}
\section{Introduction}
This paper focuses on the safety verification problem of trajectories of the states of autonomous systems in the presence of uncertainties.
Trajectory planners such as rapidly-exploring random tree (RRT), probabilistic roadmap (PRM), trajectory-optimization based planners, virtual potential field methods, and deep-learning based planners are commonly used to generate safe, e.g., obstacle-free, trajectories \cite{lavalle2006planning,lynch2017modern,reif1994motion,aradi2020survey}. However, they usually neglect the planning and execution uncertainties such as sensor and perception noises and external and control disturbances; Hence, the safety of the generated plans can not be guaranteed. In this paper, we provide efficient algorithms to verify and guarantee the safety of the generated plans in the presence of uncertain nonlinear time-varying safety constraints, e.g., safety in the presence of nonlinear and non-convex uncertain moving obstacles. The provided algorithms reason about the risk to ensure the safety of the generated plans where risk is defined as the probability of not satisfying the safety constraints.

Several approaches are widely used to ensure the safety of the generated plans in the presence of probabilistic uncertainties. However, existing methods to address the safety problems either are limited to particular classes of uncertainties and safety constraints such as Gaussian linear constraints \cite{ blackmore2009convex, blackmore2010probabilistic,  schwarting2017parallel, luders2010chance} or rely on sampling based methods that need a large number of uncertainty samples, e.g., Monte Carlo simulation,
\cite{cannon2017chance,  calafiore2006scenario,  janson2018monte, althoff2012safety}. Such algorithms are not suitable for online planning problems and can not assure the safety in the presence of nonlinear safety constraints and non-Gaussian uncertainties. To ensure safety in the presence of probabilistic nonlinear safety constraints, we in \cite{Contour,Risk_Ind} provide polynomial optimization based methods. Provided methods use convex optimizations in the form of semidefinite programs to compute the risk. However, such optimization-based methods are limited to the small state-spaces and are not suitable for online computations.

 In this paper, we provide novel convex algorithms to verify the safety of the planned trajectories in the presence of 
 probabilistic nonlinear time-varying safety constraints e.g., environments that contain obstacles with probabilistic location, size, geometry, and trajectories with arbitrary probabilistic distributions. The provided approaches verify the safety of the given trajectory and its neighborhood to account for the execution uncertainties and risk. 
 
 In the proposed approaches, we first use the moments of the probability distributions of the uncertainties to transform the probabilistic safety constraints to a set of deterministic safety constraints. We then use convex methods based on the nonnegativity conditions of the polynomials, i.e., sum-of-squares polynomials, to verify the obtained deterministic safety constraints over the entire planning time horizon without the need for time discretization.
 The complexity of the provided methods is independent of the size of the planning time horizon; Hence, the safety verification problems over long planning horizons can be easily addressed.
 
The outline of the paper is as follows: in Section II, the notation adopted in the paper and definitions on polynomials
and moments are presented. In Section III, we define the risk-aware safety verification problems. In Section IV and V, we provide the convex algorithms to verify the safety of the given trajectory and its neighborhood. In Section VI, to illustrate the performance of the proposed methods, we present experimental results on safety verification problems of self-driving vehicles and autonomous aerial vehicles. Finally, concluding remarks are given in Section VII.

\section{Notation and Definitions} \label{sec_def}

Given a vector $\mathbf{x}\in\R^{n_x}$ and multi-index $\alpha\in\mathbb{N}^{n_x}$, let $x^\alpha = \prod_{i=1}^nx_i^{\alpha_i}$. Also, given $\mathbf{x}^*\in \mathbb{R}^{n_x}$, let $\mathcal{Q}(\mathbf{x}^*)$ be the hyper-ellipsoid centered at $\mathbf{x}^*$ with positive semidefinite matrix $Q \in \mathbb{R}^{n_x \times n_x}$ as  $\mathcal{Q}(\mathbf{x}^*) =\{ \mathbf{x} \in \mathbb{R}^{n_x}:  (\mathbf{x}-\mathbf{x}^*)^T{Q}(\mathbf{x}-\mathbf{x}^*)\leq 1 \}$.


\textbf{Polynomials:} Given polynomial
{\small$\mathcal{P}(\mathbf{x}):\mathbb{R}^{n_x}\rightarrow\mathbb{R}$}, we represent {\small $\mathcal{P}$} as {\small $\sum_{\alpha\in\mathbb{N}^{n_x}} p_\alpha x^\alpha$} where 
{\small $\{x^\alpha\}_{\alpha\in \mathbb{N}^{n_x}}$} are the standard monomial basis, {\small $\mathbf{p}=\{p_\alpha\}_{\alpha\in\mathbb{N}^{n_x}}$} denotes the coefficients, and $\alpha \in \mathbb N ^{n_x}$. In this paper, we use polynomials to represent the uncertain safety constraints and continuous-time trajectories. For example the set \begin{small}$ \mathcal{X}_{obs}(\omega_1,\omega_2,t)=\lbrace (x_1,x_2): \left(x_1-p_{x_1}(t,\omega_1)\right)^2+\left(x_2-p_{x_2}(t,\omega_2)\right)^2-1 \leq 0 \rbrace$\end{small} represents a moving circle-shaped obstacle with uncertain polynomial trajectories $p_{x_1}(t,\omega_1)$ and $ p_{x_2}(t,\omega_2)$ where $\omega_1$ and $\omega_2$ are random variables. Also, 
\begin{small}$\left[\begin{matrix}
x_1(t)\\
x_2(t)
\end{matrix}\right]=\left[\begin{matrix}
1\\
-1\end{matrix}\right]t + \left[\begin{matrix}
0.5\\
2\end{matrix}\right]t^2$\end{small}, $t\in [0,1]$ is an example of polynomial trajectory of order 2 in 2-dimensional state-space between the states $\mathbf{x}(0)=(0,0)$ and $\mathbf{x}(1)=(1.5,1)$.

\textbf{Moments of Probability Distributions:}
Moments of random variables are the generalization of mean and covariance and are defined as
expected values of the monomials of random variables. More precisely, given $\alpha\in\mathbb{N}^{n_x}$ where $\alpha=\sum_{i=1}^{n_x}\alpha_i$, moment of order $\alpha$ of random vector $\mathbf{\omega}$ is defined as $\mathbb{E}[ \Pi_{i=1}^{n_x} \omega_i^{\alpha_i}]$. For example, sequence of the moments of order $\alpha=2$ for $n_x=2$ is defined as 
\begin{small}$\left[\mathbb{E}[w_1^{2}], \mathbb{E}[\omega_1\omega_2],\mathbb{E}[\omega_2^2] \right]$\end{small}. 
Moments of random vectors can be easily computed using the characteristic function of the probability distributions \cite{MOM1}. We will use  finite sequence of the moments to represent non-Gaussian probability distributions.

\textbf{Sum of Squares Polynomials:} 
Polynomial {\small$\mathcal{P}(x)$} is a sum of squares polynomial if it can be written as a sum of \emph{finitely} many squared polynomials, i.e., {\small$\mathcal{P}(x)= \sum_{j=1}^{m} h_j(x)^2$} for some $m<\infty$ and polynomials $h_j(x)$ for $1\leq j\leq m$. 
SOS condition, i.e., {\small$\mathcal{P}(x) \in SOS$}, can be represented as a convex constraint of the form of a linear matrix inequality (LMI) in terms of the coefficients of the polynomial, i.e.,
{\small $\mathcal{P}(x) \in SOS \Leftrightarrow \mathcal{P}(x)=\mathbf{x}^TA\mathbf{x}$}, where $\mathbf{x}$ is the vector of standard basis and $A$ is a positive semidefinite matrix in terms of the coefficients of the polynomial \cite{SOS1,SOS2,SOS3}. One can use the packages like Yalmip \cite{Yalmip_1} and Spotless \cite{Spot_1} to check the SOS condition of the polynomials. In this paper, we will use SOS conditions to describe the risk bounded safety conditions of the state trajectories of the autonomous systems.

\section{Problem Formulation}

Suppose $\mathcal{X}\in \mathbb{R}^{n_x}$ is the state-space and sets $\mathcal{X}_{s_i}(\omega_i,t) \subset \mathcal{X}, \ i=1,...,n_{s}$ are the uncertain time-varying safe sets defined in terms of the safety constraints as follows:
\begin{equation} \label{safe_set}
\mathcal{X}_{s_i}(\omega_i,t)= \{ \mathbf{x}\in \mathcal{X}: g_i(\mathbf{x},\mathbf{\omega}_i,t) \geq 0  \}, \ i=1,...,{n_{s}}
\end{equation}
 where $g_i:  \mathbb{R}^{n_x+n_{\omega}+1} \rightarrow \mathbb{R}, i=1,...,n_{s}$ are the given polynomials that describe the safety constraints and $\omega_i \in \mathbb{R}^{n_{\omega}}, i=1,...,n_{s}$ are the uncertain parameters with known probability distributions. The safe sets in \eqref{safe_set} are, in general, \textit{nonconvex}. Also, let $\mathcal{P}(t): [t_0, t_f] \rightarrow \mathbb{R}^{n_x}$ be the given continuous-time state polynomial trajectory generated by the planner over the planning time horizon $t\in[t_0,t_f]$ between the start and final states $\mathbf{x}_0$ and $\mathbf{x}_f$, i.e., $\mathbf{x}(t)=\mathcal{P}(t), \mathcal{P}(t_0)=\mathbf{x}_0, \mathcal{P}(t_f)=\mathbf{x}_f$. \\

 \textbf{Risk-Aware Safety Verification Problem:}
 Given the uncertain safety constraints in \eqref{safe_set} and the planned state trajectory $\mathcal{P}(t)$, we define the risk at time $t$ as the probability of violation of the uncertain safety constraints by the given trajectory $\mathcal{P}(t)$  at time $t$. 
 In the risk-aware safety verification problem, we want to make sure that the probability of violation of the uncertain safety constraints, i.e., risk, is bounded over the entire planning time horizon $[t_0, t_f]$. More precisely, we aim at verifying the following probabilistic safety constraints:
\begin{equation} \label{cc_1}
    \hbox{Prob}\left( \mathcal{P}(t) \notin \mathcal{X}_{s_i}(\omega_i,t)    \right) \leq \Delta, \ \forall t\in [t_0,t_f] \ |_{i=1}^{n_s}
\end{equation}
where $ 0 \leq \Delta \in \mathbb{R} \leq 1 $ is the given acceptable risk level.\\

\begin{figure}
	\centering
	\includegraphics[scale=0.32]{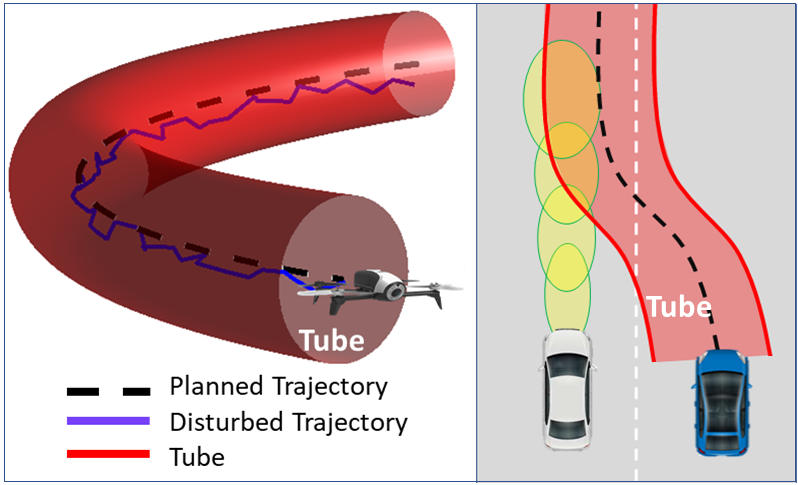}
	\caption{{\footnotesize Tube around the planned trajectory. In the tube-based safety verification problem, we want to make sure that tube satisfies the probabilistic safety constraints.}}
	\label{fig_01}
\end{figure}

\textbf{Tube-Based Risk-Aware Safety Verification Problem:} 
Due to the external disturbances, states of the autonomous systems tend to deviate from the planned trajectory. Hence, to ensure the safety, we want to make sure that all the trajectories in the neighborhood of the planned trajectory are also safe. 
For this purpose, we use tubes to represent the family of the trajectories in the neighborhood of the planned trajectory.
More precisely, we use hyper-ellipsoids defined around the given trajectory $\mathcal{P}(t)$, e.g., Figure \ref{fig_01}, to model the tube as follows:
\begin{equation}\label{tube}
\mathcal{Q}(\mathcal{P}(t)) =\{ \mathbf{x} \in \mathbb{R}^{n_x}:  (\mathbf{x}-\mathcal{P}(t))^T{Q}(\mathbf{x}-\mathcal{P}(t))\leq 1 \}   
\end{equation}
where $Q\in \mathbb{R}^{n_x \times n_x}$ is the given positive definite matrix and $t\in[t_0,t_f]$. 
To assure the safety of the family of the trajectories inside the tube, we need to make sure that the defined tube in \eqref{tube} is safe. At time $t$, the defined tube is safe, if it is a subset of the given safe sets, i.e., $\mathcal{Q}(\mathcal{P}(t)) \subset \mathcal{X}_{s_i}, i=1,...,n_s$. 

In  the  tube-based risk-aware safety verification problem, we want to make sure that the probability of violation of the uncertain safety constraints by the given tube $\mathcal{Q}(\mathcal{P}(t))$ is bounded. More precisely, we aim at verifying the following probabilistic safety constraints:
\begin{equation} \label{cc_t1}
    \hbox{Prob}\left( \mathcal{Q}(\mathcal{P}(t)) \not\subset \mathcal{X}_{s_i}(\omega_i,t)    \right) \leq \Delta, \ \forall t\in [t_0,t_f] \ |_{i=1}^{n_s}
\end{equation}
where $ 0 \leq \Delta \in \mathbb{R} \leq 1 $ is the given acceptable risk level. Note that if the given tube satisfies the probabilistic safety constraints in \eqref{cc_t1}, then any trajectory inside the tube also satisfies the probabilistic safety constraints.

Evaluating the probabilistic safety constraints in \eqref{cc_1} and \eqref{cc_t1} are challenging and hard, because we need to deal with multivariate integrals of the probabilistic constraints defined over the nonlinear and nonconvex safe sets. Moreover, we need to evaluate the probabilistic safety constraints over the entire continuous planning time horizon $[t_0,t_f]$.
In this paper, we provide a fast convex algorithm to efficiently evaluate the probabilistic safety constraints in the presence of nonlinear and nonconvex safe sets, arbitrary probability distributions, and long planning horizons, in real-time, without the need for uncertainty samples and time discretization.

\textit{\textbf{Remark 1}}: Probabilistic safety constraints in \eqref{cc_1} and \eqref{cc_t1} can be easily incorporated into standard motion planning algorithms 
such as rapidly-exploring random tree (RRT), probabilistic roadmap (PRM), motion primitive based algorithms, and deep-learning based planners, to verify the safety of the generated plans of autonomous systems in the presence of uncertainties.





\section{Risk-Aware Safety  Verification}\label{sec_traj_safe}

In this section, we address the risk-aware safety verification problem involving the probabilistic safety constraints in \eqref{cc_1}. For this purpose, we will first transform the probabilistic safety verification problem into a deterministic safety verification problem involving deterministic safety constraints. Next, we provide a convex method based on sum-of-squares polynomials to verify the safety of the given trajectory over the entire planning time horizon with respect to the obtained deterministic safety constraints without the need for time discretization.

\subsection{Deterministic Safety Constraints}

Given the uncertain safe sets in \eqref{safe_set} and probabilistic safety constraints in \eqref{cc_1}, in this section, we look for the set of all states $\mathbf{x}\in\mathcal{X}$ that satisfies the probabilistic safety constraints of the form $\hbox{Prob}\left( \mathbf{x} \notin \mathcal{X}_{s_i}(\omega_i,t)    \right) \leq \Delta, i=1,...,n_s $. By doing so, to assure the safety, we just need to make sure that the given trajectory is inside the obtained safe set of the states.
More precisely, we aim at finding the following sets\footnote{In \cite{Contour, Contour22} , we use "$\Delta$-risk contours" to refer to $\mathcal{C}_{r}^{\Delta}$ sets in \eqref{cc_3} }:
\begin{equation} \label{cc_3}
  \mathcal{C}_{r_i}^{\Delta}(t)= \{\mathbf{x} \in \mathcal{X}: \hbox{Prob}\left( \mathbf{x} \notin \mathcal{X}_{s_i}(\omega_i,t)    \right) \leq \Delta \}, \ |_{i=1}^{n_s}
\end{equation}
Any state trajectory $\mathbf{x}(t) \in \mathcal{C}_{r_i}^{\Delta}(t) |_{i=1}^{n_s} \forall t\in[t_0,t_f]$
satisfies the probabilistic safety constraints in \eqref{cc_1}. The main idea to construct the sets $\mathcal{C}_{r_i}^{\Delta}(t),i=1,...,n_s$ in \eqref{cc_3} is to replace the probabilistic constraints, i.e., $\hbox{Prob}\left( \mathbf{x} \notin \mathcal{X}_{s_i}(\omega_i,t)    \right) \leq \Delta,i=1,...,n_s$, with deterministic constraints in terms of $\mathbf{x}$ and time $t$. In this paper, we provide an optimization free approach as follows \cite{Contour22}:




Let $P_{1i}(\mathbf{x},t), P_{2i}(\mathbf{x},t), i=1,...,n_s$ be the polynomials defined in terms of the polynomials of the uncertain safe sets in \eqref{safe_set} as follows:
\begin{small}
\begin{equation}\label{poly_det_1}
P_{1i}(\mathbf{x},t)=\mathbb{E}[g^2_i(\mathbf{x},\omega_i,t)], \ 
P_{2i}(\mathbf{x},t)=\mathbb{E}[g_i(\mathbf{x},\omega_i,t)] \ |_{i=1}^{n_s}
\end{equation}
\end{small}
where the expectation is taken with respect to the distribution of uncertainties $\omega_i$. We can compute polynomials $P_{1i}(\mathbf{x},t)$, $ P_{2i}(\mathbf{x},t)$, $i=1,...,n_s$ in terms of $\mathbf{x}$, $t$, and known moments of uncertain parameters $\omega_i$. More precisely, $P_{1i}(\mathbf{x},t), P_{2i}(\mathbf{x},t), i=1,...,n_s$ are polynomials in $\mathbf{x}$ and $t$ whose coefficients are in terms of the moments of $\omega_i$ and the coefficients of the polynomials of the uncertain safe sets in \eqref{safe_set}.

Also, let  $\hat{\mathcal{C}}_{r_i}^{\Delta}(t), i=1,...,n_s$ be the sets defined in terms of the polynomials 
$P_{1i}(\mathbf{x},t), P_{2i}(\mathbf{x},t), i=1,...,n_s$ as follows:
\begin{equation} \label{cc_det_1}
  \hat{\mathcal{C}}_{r_i}^{\Delta}(t)=  \left\lbrace \mathbf{x} \in \mathcal{X}:
  \begin{array}{cc}
  \frac{P_{1i}(\mathbf{x},t)-P^2_{2i}(\mathbf{x},t)}{P_{1i}(\mathbf{x},t)} \leq \Delta, \\
  P_{2i}(\mathbf{x},t) \geq 0
  \end{array}
   \right\rbrace \  |_{i=1}^{n_s}
\end{equation}
The following result holds true.\\

\textbf{Theorem 1:} The sets $\hat{\mathcal{C}}_{r_i}^{\Delta}(t), i=1,...,n_s$ in \eqref{cc_det_1} are the inner approximations of the set of all safe states ${\mathcal{C}}_{r_i}^{\Delta}(t), i=1,...,n_s$ in \eqref{cc_3}.

\textit{Proof}: Appendix A.  \hfill $\blacksquare$

We can interpret the obtained results in \eqref{cc_det_1} as
follows. For any states $\mathbf{x}\in \mathcal{X}$, the rational polynomial $\frac{P_{1i}(\mathbf{x},t)-P^2_{2i}(\mathbf{x},t)}{P_{1i}(\mathbf{x},t)}$ is the upper bound of the probability $\hbox{Prob}\left( \mathbf{x} \notin \mathcal{X}_{s_i}(\omega_i,t)    \right)$ at time $t$, if the expectation of being safe at time $t$ is nonnegative, i.e., $P_{2i}(\mathbf{x},t)=\mathbb{E}[g_i(\mathbf{x},\omega_i,t)] \geq 0$. Hence, the sets in \eqref{cc_det_1} describe the inner approximations of the sets of all safe states in \eqref{cc_3}.

Note that since $\hat{\mathcal{C}}^{\Delta}_{r_i}(t), i=1,...,n_s$ are inner approximations of ${\mathcal{C}}^{\Delta}_{r_i}(t)$, any trajectory $\mathbf{x}(t) \in \hat{\mathcal{C}}^{\Delta}_{r_i}(t), i=1,...,n_s \ \forall t\in [t_0,t_f]$ is guaranteed to have a risk less or equal to $\Delta$. Hence, we can use the obtained deterministic safety constraints in $\hat{\mathcal{C}}^{\Delta}_{r_i}(t), i=1,...,n_s$ to ensure the safety of the given state trajectory. We now provide an illustrative example to show the performance of the proposed method to construct the set of safe states $\hat{\mathcal{C}}^{\Delta}_{r_i}(t), i=1,...,n_s$.\\

\textbf{Illustrative Example 1:}
Consider the following illustrative example where $\mathcal{X}=[-1,1]^2$. 
In this example, we consider 2 cases including time-invariant  and time-varying probabilistic safe sets.\\
\textit{Case 1- Time-Invariant Probabilistic Safe Sets}: The set {\small$\mathcal{X}_{s}(\omega)= \left\lbrace  (x_1,x_2) :  x_1^2+x_2^2-\omega^2 \geq 0 \right\rbrace $} represents an probabilistic safe region in the presence of circle-shaped obstacle with uncertain radius $\omega$ with uniform probability distribution over {\small$[0.3,0.4]$}, \cite{Contour}. 
Moment of order $\alpha$ of uniform distribution over $[l,u]$ is described as $\frac{u^{\alpha+1}-l^{\alpha+1}}{(u-l)(\alpha+1)}$.
To construct the deterministic safe set in \eqref{cc_det_1}, we compute polynomials \begin{small}$P_{1}(\mathbf{x}), P_{2}(\mathbf{x})$\end{small} using the polynomial of the safe set and the moments of $\omega$ as follows:
{\small$P_1(\mathbf{x})=\Exp[\left(x_1^2+x_2^2-\omega^2 \right)^2]
    =\Exp[\omega^4] - 2\Exp[\omega^2]x_1^2 - 2\Exp[\omega^2]x_2^2 + x_1^4 + 2x_1^2x_2^2 + x_2^4
    =0.01 - 0.24x_1^2 - 0.24x_2^2 + x_1^4 + 2x_1^2x_2^2 + x_2^4$} and {\small $ P_2(\mathbf{x})=\Exp[x_1^2+x_2^2-\omega^2 ]= x_1^2+x_2^2-\Exp[\omega^2] =x_1^2+x_2^2-0.12 $}. Given that safety constraints are time invariant, we drop $t$  in the notation of the polynomials.

\begin{figure}
    \centering
    \includegraphics[scale=0.12]{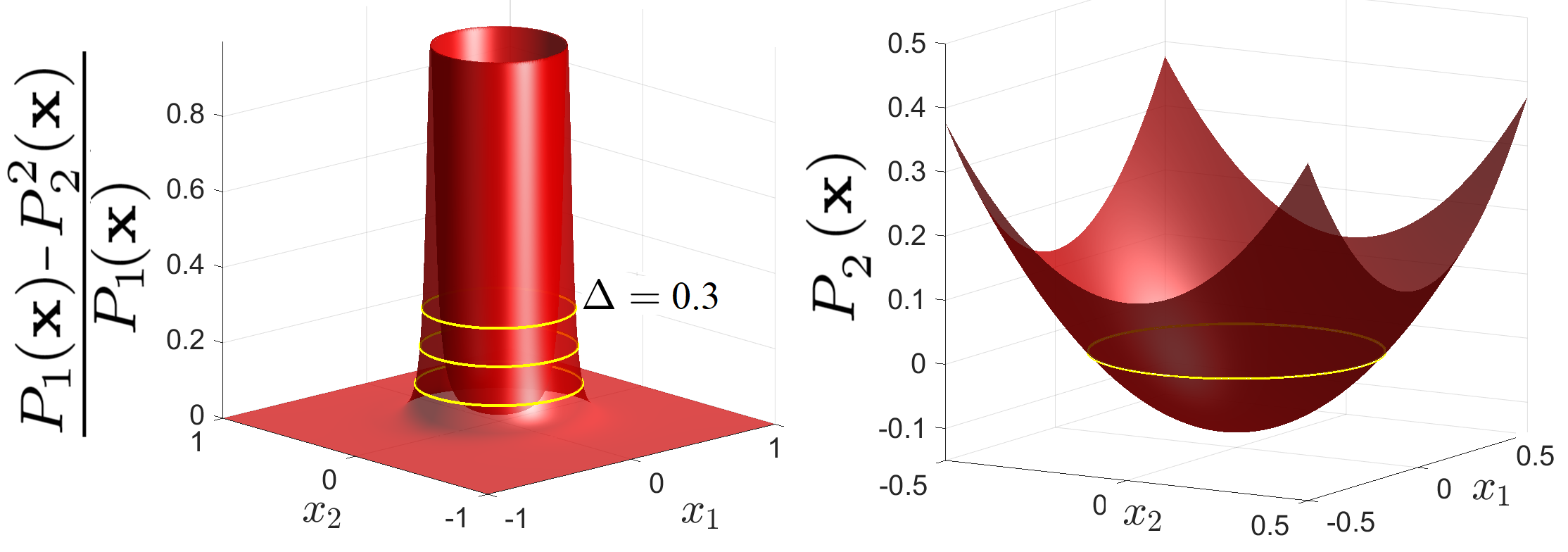}
	\caption{{\footnotesize Intersection of $\Delta$-sublevel set of rational polynomial 
	$\frac{P_{1}(\mathbf{x})-P^2_{2}(\mathbf{x})}{P_{1}(\mathbf{x})} $ and $0$-superlevel set of polynomial $P_{2}(\mathbf{x})$ constructs the deterministic safe set in \eqref{cc_det_1}.}}
    \label{fig_1}
\end{figure}
\begin{figure}
    \centering
    \includegraphics[scale=0.28]{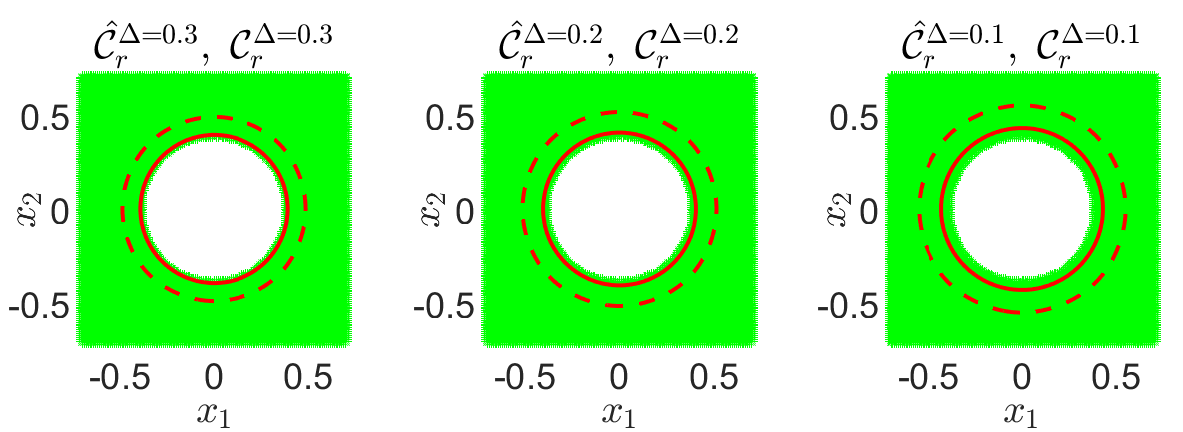}
	\caption{{\footnotesize   ${\mathcal{C}}^{\Delta}_{r}$ (green) set of all safe states described in \eqref{cc_3}  and inner approximation $\hat{\mathcal{C}}^{\Delta}_{r}$ obtained using i) our proposed method described in Eq.  \eqref{cc_det_1} (outside of the solid-line) and ii) the proposed optimization based method in (\cite{Contour}, Fig.4) (outside of the dashed-line).
	Probability of collision with the uncertain obstacle inside the set $\hat{\mathcal{C}}^{\Delta}_{r}$ is less or equal to $\Delta$.
	}}
    \label{fig_2}
\end{figure}

As shown in Figure \ref{fig_1}, we use the level sets of the rational polynomial \begin{small}$\frac{P_{1}(\mathbf{x})-P^2_{2}(\mathbf{x})}{P_{1}(\mathbf{x})}$\end{small} and polynomial \begin{small}$P_{2}(\mathbf{x})$\end{small} to construct the inner approximation of the deterministic safe set described in \eqref{cc_det_1}. We construct the safe sets for different risk levels $\Delta=[0.5,0.3,0.1]$ as shown in Figure \ref{fig_2}. We also compare our proposed method with the optimization based method in \cite{Contour} as shown in Figure \ref{fig_2}. Our proposed method obtains the tight inner approximations of the set of all safe states ${\mathcal{C}}_{r_i}^{\Delta}(t), i=1,...,n_s$ and it is suitable for online large-scale planning problems. For more information see Illustrative Example 1 in \cite{Contour22}.

\begin{figure}[h]
    \centering
    \includegraphics[scale=0.25]{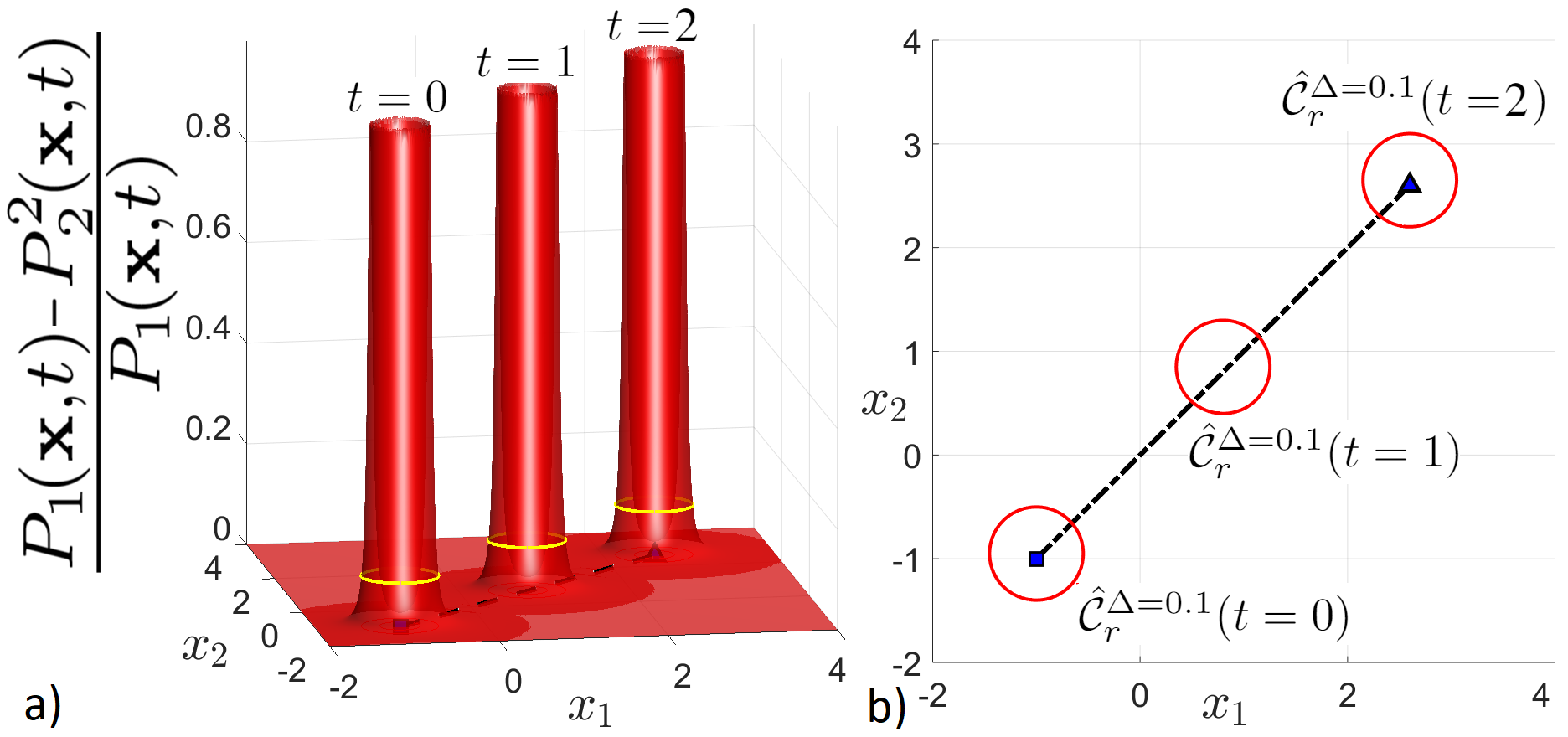}
	\caption{{\footnotesize 
	a) Rational polynomial $\frac{P_{1}(\mathbf{x},t)-P^2_{2}(\mathbf{x},t)}{P_{1}(\mathbf{x},t)}$  at time steps $t=0,1,2$, b) Time-varying deterministic safe sets for $\Delta=0.1$ at time steps $t=0,1,2$ 
	described in \eqref{cc_det_1}, (outside of the solid-line). Dashed line shows the expected value of the given uncertain trajectory, i.e., $\Exp[(p_{x_1}(t,\omega_2),p_{x_2}(t,\omega_3))]$. At each time $t$, probability of collision with the moving uncertain obstacle inside the $\hat{\mathcal{C}}^{\Delta}_{r}(t)$ is less or equal to $\Delta=0.1$. 
	}}
    \label{fig_3}
\end{figure}

\textit{Case 2- Time-Varying Safe Set:} 
The set {\small$\mathcal{X}_{s}(\omega,t)= \left\lbrace  (x_1,x_2) :  (x_1-p_{x_1}(t,\omega_2))^2+(x_2-p_{x_2}(t,\omega_3))^2-\omega_1^2 \geq 0 \right\rbrace$}  represents a time-varying probabilistic safe set in the presence of moving circle-shaped obstacle with uncertain radius $\omega_1$ and uncertain trajectories \begin{small}$ p_{x_1}(t,\omega_2)=1.8t-1+0.2\omega_2,\  p_{x_2}(t,\omega_3)=1.8t-1+0.1\omega_3 $\end{small} that describe the uncertain motion of the obstacle over the time horizon $t\in[0,2]$. Uncertain parameters have uniform, normal and, Beta distributions as \begin{small}$\omega_1 \sim Uniform[0.3,0.4]$, $\omega_2 \sim \mathcal{N}(0,0.1)$, $\omega_3 \sim Beta(3,3)$\end{small}. Similar to case 1, we compute the rational polynomial \begin{small}$\frac{P_{1}(\mathbf{x},t)-P^2_{2}(\mathbf{x},t)}{P_{1}(\mathbf{x},t)}$\end{small} and \begin{small}$P_{2}(\mathbf{x},t)$\end{small} using the moments of uncertainties $\omega_i, i=1,...3$ and the polynomial of the safe set.
We then use \eqref{cc_det_1} to construct the time-varying deterministic safe set.
Figure \ref{fig_3} shows the obtained safe set for $\Delta=0.1$ at time steps $t=0,1,2$ along the given uncertain trajectory \begin{small}$\left( p_{x_1}(t,\omega_2),\  p_{x_2}(t,\omega_3) \right)$\end{small}.\\

\subsection{Continuous-Time Safety Verification}
In this section, we use the obtained
deterministic safety constraints in $\hat{\mathcal{C}}^{\Delta}_{r_i}(t), i=1,...,n_s$ to verify the safety of the given state trajectory over the entire planning time horizon as follows:

Let $\mathbf{x}(t)=\mathcal{P}(t): [t_0, t_f] \rightarrow \mathbb{R}^{n_x}$ be the given trajectory of the states over the planning time horizon $t\in[t_0,t_f]$ between the start and final states $\mathbf{x}_0$ and $\mathbf{x}_f$.
Also, let $P_{1i}(\mathbf{x},t), P_{2i}(\mathbf{x},t), i=1,...,n_s$ be the polynomials of the obtained deterministic safety constraints in $\hat{\mathcal{C}}^{\Delta}_{r_i}(t), i=1,...,n_s$.

The following result holds true.\\
 
\textbf{ Theorem 2}: The planned state trajectory $\mathcal{P}(t)$ satisfies the probabilistic safety constraints in \eqref{cc_1} over the entire planning time horizon $t\in[t_0,t_f]$ if the polynomials of the deterministic safety constraints $\hat{\mathcal{C}}^{\Delta}_{r_i}(t), i=1,...,n_s$ in \eqref{cc_det_1} take the following SOS representation:
\begin{align}\label{SOS_Cond1}
    P^2_{2i}\left(\mathcal{P}(t),t\right)-& (1-\Delta)P_{1i}(\mathcal{P}(t),t)\\
   &= {\sigma_0}_i(t)+ {\sigma_1}_i(t)(t-t_0)(t_f-t) \ |_{i=1}^{n_s} \notag
\end{align}
\begin{align}\label{SOS_Cond2}
    P_{2i}(\mathcal{P}(t),t)={\sigma_2}_i(t)+ {\sigma_3}_i(t)(t-t_0)(t_f-t) \ |_{i=1}^{n_s} 
\end{align}
where ${\sigma_0}_i(t), {\sigma_1}_i(t), {\sigma_2}_i(t),{\sigma_3}_i(t) i=1,...,n_s$ are SOS polynomials with appropriate degrees \cite{putinar1993positive,SOS2,SOS3}. \\
\textit{Proof}: Appendix B.  \hfill $\blacksquare$

To obtain the results of Theorem 2, we use the fact that any trajectory $\mathcal{P}(t) \in \hat{\mathcal{C}}_{{r}_i}^{\Delta}(t) |_{i=1}^{n_s} \  \forall t\in[t_0,t_f]$ satisfies the probabilistic safety constraints in \eqref{cc_1}. Also, we use the nonnegativity conditions of the polynomials of $\hat{\mathcal{C}}^{\Delta}_{r_i}(t), i=1,...,n_s$ over the given time interval \begin{small}$\{ t: (t-t_0)(t_f-t) \geq 0 \}$\end{small}. Note that SOS conditions in \eqref{SOS_Cond1} and \eqref{SOS_Cond2} are convex and can be easily verified using the packages like Yalmip \cite{Yalmip_1} and Spotless \cite{Spot_1}. We   now   provide   an   illustrative   example   to show  the  performance  of  the  proposed  method.\\

\begin{figure}[t]
    \centering
    \includegraphics[scale=0.245]{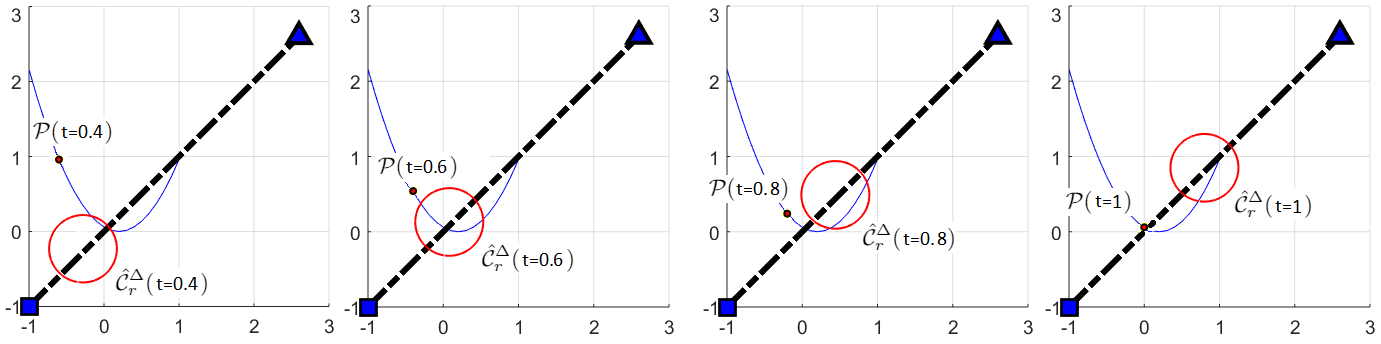}
	\caption{{\footnotesize 
	 Illustrative Example 2: Given state trajectory (blue solid-line), time-varying deterministic safe sets $\hat{\mathcal{C}}^{\Delta=0.1}_{r}(t)$ (outside of the red curve) at time steps $t=0.4,0.6,0.8, 1$. At each time step $t$, the given trajectory is inside the safe set $\hat{\mathcal{C}}^{\Delta=0.1}_{r}(t)$. Hence, it satisfies the probabilistic safety constraints in \eqref{cc_1}. 
	}}
    \label{fig_ill4}
\end{figure}

\textbf{Illustrative Example 2:} Consider the time-varying safety constraint in illustrative Example 1 Case 2. In this example, we want to verify the safety of the given state trajectory $x_1(t)=t-1$, $x_2(t)=1.5(t-1.2)^2$ over the planning horizon $t\in [0,2]$
with respect to the probabilistic constraints in \eqref{cc_1} with $\Delta=0.1$ in the presence of the moving uncertain obstacle. Using Spotless, we verify that the given state trajectory satisfies the conditions in \eqref{SOS_Cond1} and \eqref{SOS_Cond2}. Hence, it satisfies the risk bounded safety constraints over the entire planning time horizon $t\in [0,2]$. Figure \ref{fig_ill4} shows the given state trajectory and time-varying deterministic safe sets $\hat{\mathcal{C}}^{\Delta=0.1}_{r}(t)$ at time steps $t=0.4,0.6,0.8, 1$. In Illustrative Example 3  in \cite{Contour22}, we compare the SOS-based safety verification method with Monte Carlo-based method and show that 
the SOS-based method not only is faster but also provide safety guarantees.

\section{Tube-Based Risk-Aware Safety  Verification}

In  this  section,  we  address  the  tube-based risk-aware  safety  verification  problem to verify the safety of the given tube $\mathcal{Q}(\mathcal{P}(t))$ in \eqref{tube} with respect to the probabilistic safety constraints  in
\eqref{cc_t1}. For this purpose, we use the obtained deterministic safety constraints $\hat{\mathcal{C}}^{\Delta}_{r_i}(t), i=1,...,n_s$ in \eqref{cc_det_1} and  sum-of-squares conditions of the polynomials as follows:\\

\textbf{ Theorem 3}: The given tube $\mathcal{Q}(\mathcal{P}(t))$ satisfies the probabilistic safety constraints in \eqref{cc_t1} over the entire planning time horizon $t\in[t_0,t_f]$ if the polynomials of the deterministic safety constraints $\hat{\mathcal{C}}^{\Delta}_{r_i}(t), i=1,...,n_s$ in \eqref{cc_det_1} take the following SOS representation:
\begin{small}
\begin{align}\label{SOS_t_Cond1}
    & P^2_{2i}  \left(\mathcal{P}(t)+\hat{\mathbf{x}}_0,t\right)- (1-\Delta)P_{1i}(\mathcal{P}(t)+\hat{\mathbf{x}}_0,t)=\\
   & {\sigma_0}_i(t,\hat{\mathbf{x}}_0)+ {\sigma_1}_i(t,\hat{\mathbf{x}}_0)(t-t_0)(t_f-t) + {\sigma_2}_i(t,\hat{\mathbf{x}}_0)(1-\hat{\mathbf{x}}_0^T{Q}\hat{\mathbf{x}}_0) |_{i=1}^{n_s} \notag
\end{align}\end{small}
\begin{small}\begin{align}\label{SOS_t_Cond2}
    & P_{2i} (\mathcal{P}(t)+\hat{\mathbf{x}}_0,t)=\\
    & {\sigma_3}_i(t,\hat{\mathbf{x}}_0)+ {\sigma_4}_i(t,\hat{\mathbf{x}}_0)(t-t_0)(t_f-t) + {\sigma_5}_i(t,\hat{\mathbf{x}}_0)(1-\hat{\mathbf{x}}_0^T{Q}\hat{\mathbf{x}}_0) |_{i=1}^{n_s}  \notag
\end{align}\end{small}where $\hat{\mathbf{x}}_0 \in \mathbb{R}^{n_x}$ is the variable vector, ${\sigma_j}_i(t,\hat{\mathbf{x}}_0),j=0,...,5, i=1,...,n_s$ are SOS polynomials with appropriate degrees and
$Q\in \mathbb{R}^{n_x \times n_x}$ is the given positive definite matrix in \eqref{tube}. \\
\textit{Proof}: Appendix C.  \hfill $\blacksquare$

To obtain the results of Theorem 3, we use 
the fact that any tube $\mathcal{Q}(\mathcal{P}(t)) \subset \hat{\mathcal{C}}_{{r}_i}^{\Delta}(t) |_{i=1}^{n_s} \  \forall t\in[t_0,t_f]$ satisfies the probabilistic safety constraints in \eqref{cc_t1}. Also, we use the nonnegativity conditions of the polynomials of $\hat{\mathcal{C}}^{\Delta}_{r_i}(t), i=1,...,n_s$  over the given time interval \begin{small}$\{ t: (t-t_0)(t_f-t) \geq 0 \}$\end{small} and set \begin{small}$\{ \hat{\mathbf{x}}_0: 1-\hat{\mathbf{x}}_0^T{Q}\hat{\mathbf{x}}_0 \geq 0 \}$\end{small}. Note that SOS conditions in \eqref{SOS_t_Cond1} and \eqref{SOS_t_Cond2} are convex and can be easily verified using the packages like Yalmip \cite{Yalmip_1} and Spotless \cite{Spot_1}. We   now   provide   an   illustrative   example   to show  the  performance  of  the  proposed  method.\\

\begin{figure}
    \centering
    \includegraphics[scale=0.24]{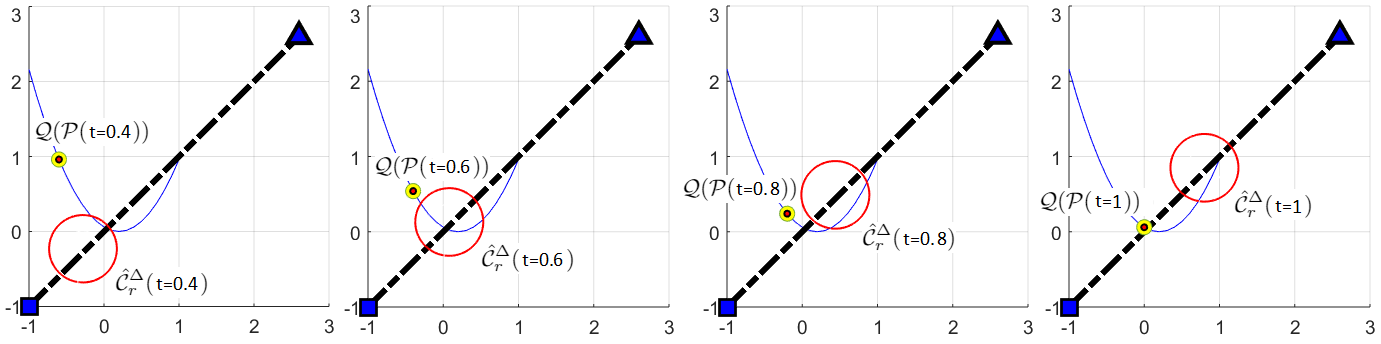}
	\caption{{\footnotesize 
	 Illustrative Example 3:
	 Given tube around the state trajectory (yellow disk), time-varying deterministic safe sets $\hat{\mathcal{C}}^{\Delta=0.1}_{r}(t)$ (outside of the red curve) at time steps $t=0.4,0.6,0.8, 1$. At each time step $t$, the given tube is a subset of the safe set $\hat{\mathcal{C}}^{\Delta=0.1}_{r}(t)$. Hence, it satisfies the probabilistic safety constraints in \eqref{cc_t1}. 
	}}
    \label{fig_ill5}
\end{figure}

\textbf{Illustrative Example 3:}
Consider the time-varying safety constraint in illustrative Example 1 Case 2 and the state trajectory of illustrative example 2. In this example, we want to verify the safety of the given tube around the planned trajectory with respect to the probabilistic constraints in \eqref{cc_t1} with $\Delta=0.1$ over the planning horizon $t\in [0,2]$. Tube is defined as
(i) disks of radius $0.1$, i.e., \begin{small}
$\mathcal{Q}(\mathcal{P}(t)) =\{ \mathbf{x} \in \mathbb{R}^{2}:  (\mathbf{x}-\mathcal{P}(t))^T\begin{bmatrix}
10 & 0\\
0 & 10
\end{bmatrix}(\mathbf{x}-\mathcal{P}(t))\leq 1 \}   $
\end{small} and 
(ii) disks of radius $0.3$. Using Spotless, we verify that the given tube in (i) satisfies the conditions in \eqref{SOS_t_Cond1} and \eqref{SOS_t_Cond2}. Hence, it satisfies the risk bounded safety constraints over the entire planning time horizon $t\in [0,2]$. 
This means that any trajectory inside the tube also satisfies the risk bounded safety constraints. Figure \ref{fig_ill5} shows the given tube and time-varying deterministic safe sets $\hat{\mathcal{C}}^{\Delta=0.1}_{r}(t)$ at time steps $t=0.4,0.6,0.8, 1$.  The given tube in (ii) does not satisfy the conditions in \eqref{SOS_t_Cond1} and \eqref{SOS_t_Cond2}. 

\textit{\textbf{Remark 2:}} The complexity of the safety SOS conditions
in Theorems 2 and 3 is independent of the size of the planning time horizon $[t_0, t_f]$ and the length of the polynomial trajectory $\mathcal{P}(t)$. Hence, they can be easily used to verify the safety of
trajectories in uncertain environments over the long planning
time horizons.

\section{Results}
In this section, several numerical examples are presented on the safety verification of autonomous systems. All computations in this section were performed on a computer with Intel i7 2.6 GHz processors and 16 GB RAM. The Spotless package \cite{Spot_1} was used to verify the SOS safety conditions\footnote{github.com/jasour/Real-Time-Risk-Bounded-Tube-based-Trajectory-Safety-Verification}.

\subsection{Autonomous Vehicle Lane-Changing Trajectory Verification}
\begin{figure}
    \centering
    \includegraphics[scale=0.2]{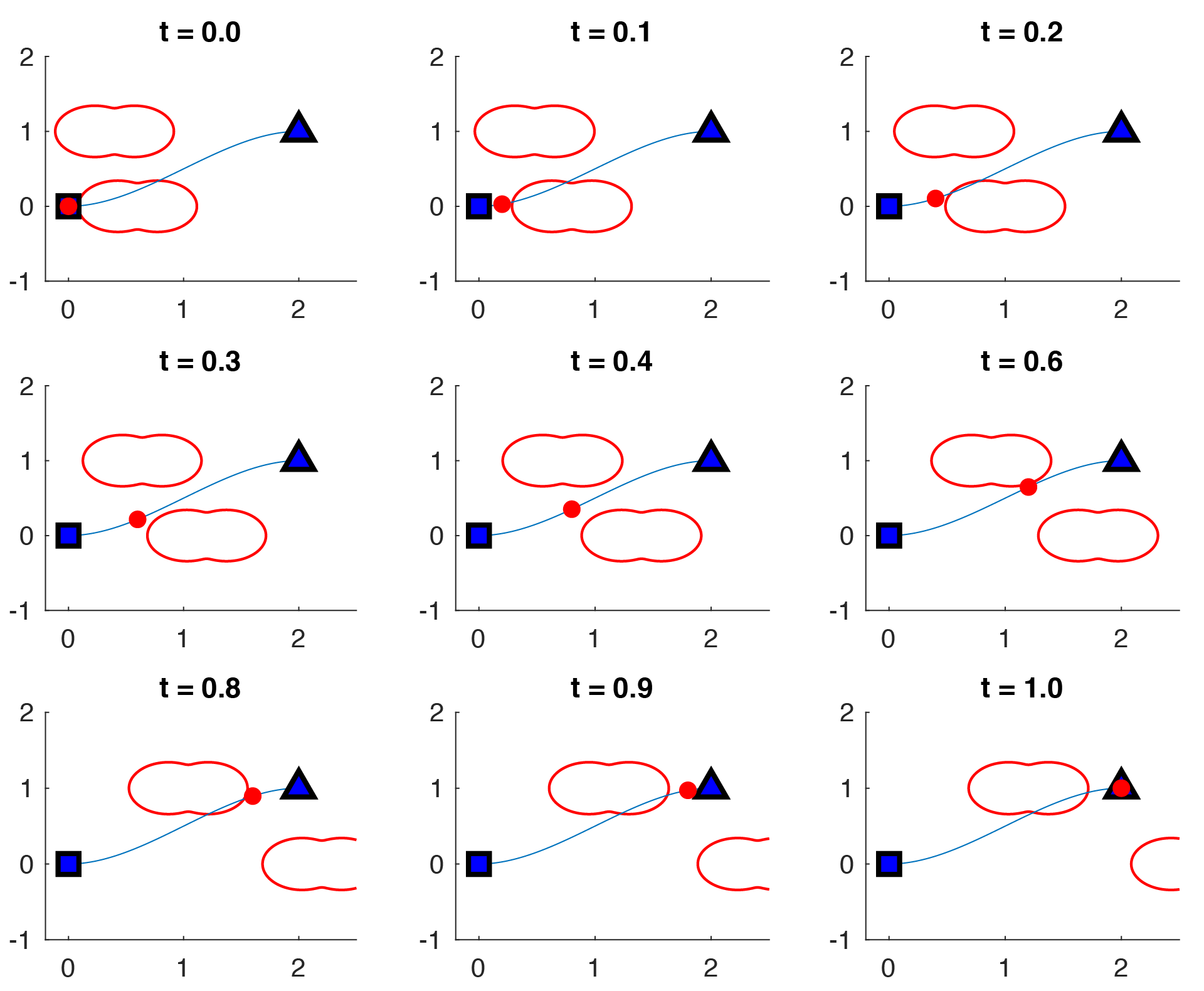}
	\caption{{\footnotesize 
        Verified risk bounded polynomial trajectory (blue curve) for lane-changing autonomous vehicle (red dot) from the starting position (square) to the goal position (triangle). The red curves represent the safe sets $\hat{\mathcal{C}}^{\Delta=0.1}_{r}(t)$ in the presence of the surrounding vehicles.
	}}
    \label{fig_3_vehicle}
\end{figure}

In this example, we verify the safety of a lane-changing autonomous vehicle's polynomial trajectory in the presence of other uncertain vehicles. The surrounding vehicles are modelled as following sets: $\mathcal{X}_{obs_1}(\omega_1,t) =  \{(x_1,x_2): 0.3^2 - (x_1-(0.4+\omega_1+0.8t))^2-(x_2-1)^2 \geq 0\}$ and $\mathcal{X}_{obs_2}(\omega_2,t) =  \{(x_1,x_2): 0.3^2 - (x_1-(0.6+\omega_2+2t))^2-x_2^2 \geq 0\}$, where $\omega_1, \omega_2 \sim Uniform[-0.1,0.1]$, i.e., two surrounding vehicles modelled as disks of radius 0.3, currently (when $t=0$) at positions $(0.4,1)$ and $(0.6,0)$, moving with velocities $(0.8,0)$ and $(2,0)$, respectively, and both having uniform uncertainty $(\omega_i,0)$ in their positions. The polynomial trajectory $x_1(t) = 2t, x_2(t) = 3t^2 - 2t^3$, $t\in [0,1]$, starting from $(0,0)$ and ending at $(2,1)$, is verified to have bounded risk of $0.1$ (Figure \ref{fig_3_vehicle}). The verification runtime is 0.45 second.
If the second surrounding vehicle's velocity reduces from $(2,0)$ to $(1,0)$, i.e., $\mathcal{X}_{obs_2}(\omega_2,t) =  \{(x_1,x_2): 0.3^2 - (x_1-(0.6+\omega_2+t))^2-x_2^2 \geq 0\}$, then the polynomial trajectory does not have the bounded risk of 0.1.

\subsection{Autonomous Vehicle Lane-Changing Tube Verification}
\begin{figure}
    \centering
    \includegraphics[scale=0.2]{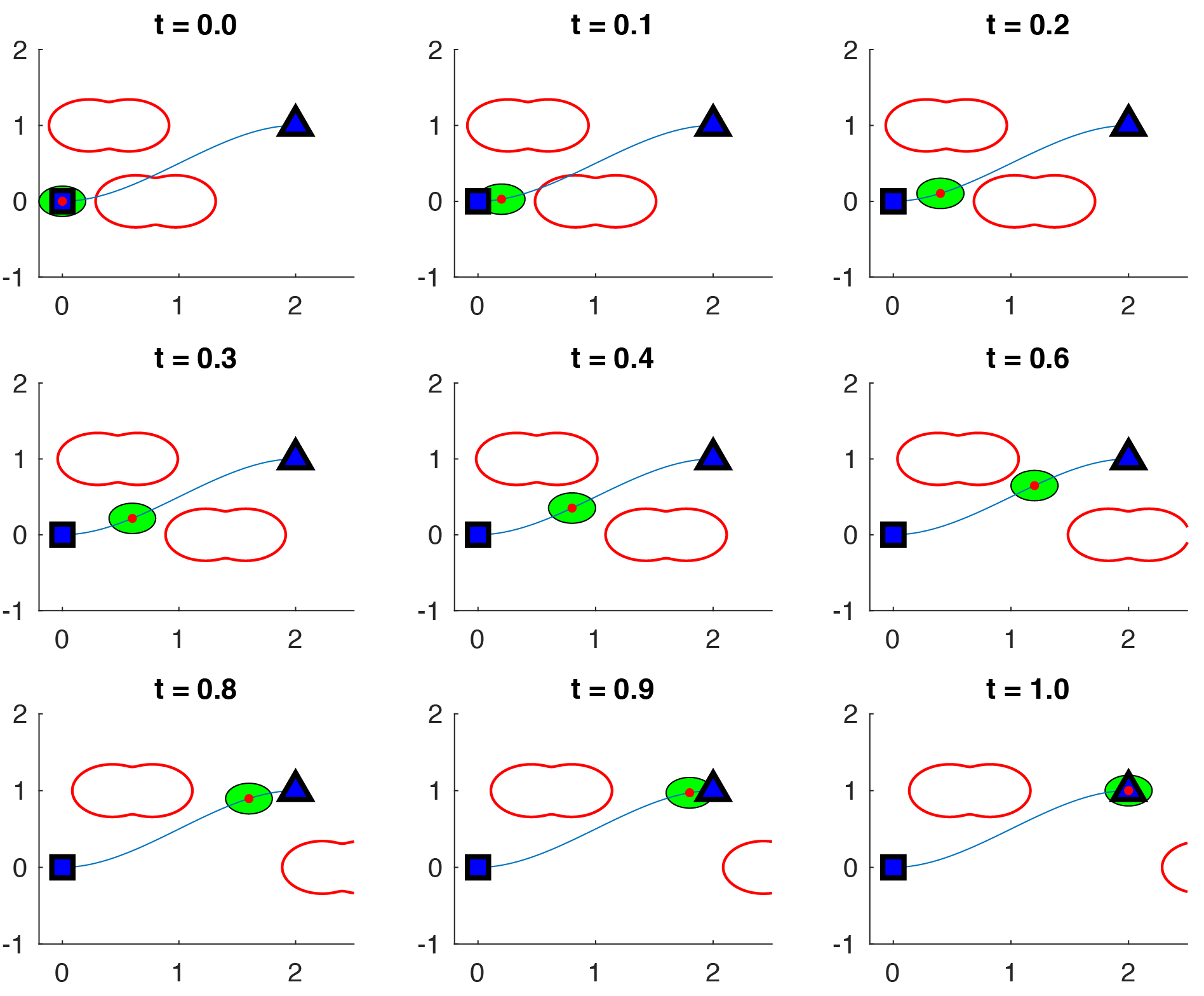}
	\caption{{\footnotesize 
        Risk bounded tube (green disk) around nominal trajectory (blue curve) for lane-changing autonomous vehicle (red dot) from the starting position (square) to the goal position (triangle). The red curves represent the safe sets $\hat{\mathcal{C}}^{\Delta=0.1}_{r}(t)$ in the presence of the surrounding vehicles.
	}}
    \label{fig_4_vehicle_tube}
\end{figure}

In this example, we verify the safety of a tube around a lane-changing autonomous vehicle's nominal trajectory in the presence of other uncertain vehicles. The surrounding vehicles are modelled as following sets: $\mathcal{X}_{obs_1}(\omega_1,t) =  \{(x_1,x_2): 0.3^2 - (x_1-(0.4+\omega_1+0.25t))^2-(x_2-1)^2 \geq 0\}$ and $\mathcal{X}_{obs_2}(\omega_2,t) =  \{(x_1,x_2): 0.3^2 - (x_1-(0.8+\omega_2+2t))^2-x_2^2 \geq 0\}$, where $\omega_1, \omega_2 \sim Uniform[-0.1,0.1]$, i.e., two surrounding vehicles modelled as disks of radius 0.3, currently (when $t=0$) at positions $(0.4,1)$ and $(0.6,0)$, moving with velocities $(0.25,0)$ and $(2,0)$, respectively, and both having uniform uncertainty $(\omega_i,0)$ in their positions.
The nominal trajectory is $x_1(t) = 2t, x_2(t) = 3t^2 - 2t^3$, $t\in [0,1]$. The state of the vehicle is assumed to be inside a disk of radius 0.2 around the nominal trajectory, and the resulting tube can be parameterized as $\{(x_1,x_2): (x_1 - 2t)^2 + (x_2 - (3t^2 - 2t^3))^2 \leq 0.1^2, t\in [0,1]\}$. The tube is verified to have bounded risk of 0.1 (Figure \ref{fig_4_vehicle_tube}). The verification runtime is 0.88 second. If the radius of the disk increases from 0.2 to 0.3, then the tube does not have bounded risk of 0.1.

\subsection{Autonomous Flight Trajectory Verification}
In this example, we verify if the trajectory of an autonomous aerial vehicle can safely pass through a cave with uncertainty on the boundary of the cave. At each time step, the free space surrounding the flight is represented by an ellipse with Gaussian noise. It is a conservative representation of the cave, because the ellipse is only part of the free space. The free space of the cave is modelled as $\mathcal{X}(\omega,t) =\{(x_1,x_2,x_3)\in\R^3:  (x_1-t+\omega_1)^2 + (x_2-(t-0.5)^2+\omega_2)^2 + (x_3-t+\omega_3)^2\leq 1 \}$, where $\omega_i \sim \mathcal{N}(\mu = 0,\sigma^2 = 0.001),i=1,2,3$ are the Gaussian noises. The polynomial trajectory $x_1(t) = t + 0.1,x_2(t) = (t-0.6)^2, x_3(t) = 1.2t - 0.1$ over the time horizon $t \in [0,1]$ is verified to have bounded risk of 0.1. The verification runtime is 0.36 second. Figure \ref{fig_5_cave_tube} shows the obtained safe set $\hat{\mathcal{C}}^{\Delta=0.1}_{r}(t)$ in the presence of the uncertain cave model.

\subsection{Autonomous Flight Tube Verification}
\begin{figure}{b}
    \centering
    \includegraphics[scale=0.14]{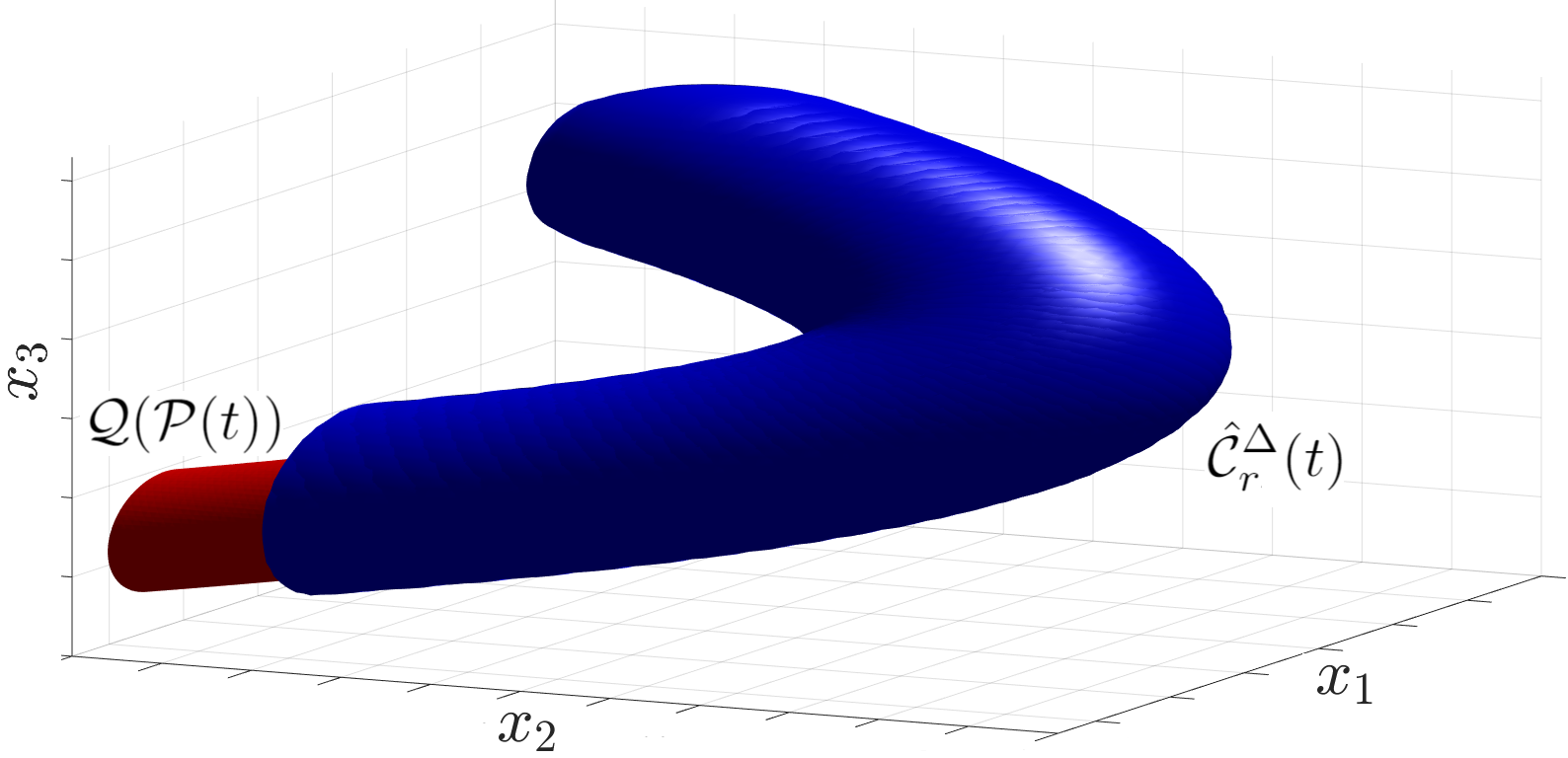}
	\caption{{\footnotesize 
The given tube $\mathcal{Q}(\mathcal{P}(t))$ around the trajectory of an autonomous aerial vehicle (red) and the obtained safe set $\hat{\mathcal{C}}^{\Delta=0.1}_{r}(t)$ in the presence of the uncertain cave model (blue). Any tube inside the blue region is guaranteed to have a bounded risk of $\Delta=0.1$.
	}}
    \label{fig_5_cave_tube}
\end{figure}

In this example, we verify if the tube around the nominal trajectory of an autonomous aerial vehicle passes safely through a cave with uncertainty on the boundary of the cave. As in the previous example, the free space of the cave is modelled as $\mathcal{X}(\omega,t) =\{(x_1,x_2,x_3)\in\R^3:  (x_1-t+\omega_1)^2 + (x_2-(t-0.5)^2+\omega_2)^2 + (x_3-t+\omega_3)^2\leq 1 \}$, where $\omega_i \sim \mathcal{N}(\mu = 0,\sigma^2 = 0.001),i=1,2,3$ are the Gaussian noise. The tube parameterized by $x_1(t) = t + 0.1 + z_1,x_2(t) = (t-0.6)^2 + z_2, x_3(t) = 1.2t - 0.1 + z_3$, $\forall (z_1,z_2,z_3) \in \{z_1^2 + z_2^2 + z_3^2 \leq 0.7^2\}$ over the time horizon $t \in [0,1]$ is verified to have bounded risk of 0.1. The verification runtime is 0.77 second.
Figure \ref{fig_5_cave_tube} shows the tube and the obtained safe set $\hat{\mathcal{C}}^{\Delta=0.1}_{r}(t)$ in the presence of the uncertain cave model.

\section{Conclusion}
In this paper, we provided fast risk-aware safety verification algorithms to verify the safety of the continuous-time trajectories in the presence of nonlinear probabilistic safety constraints. The provided algorithms  use the moments of probability distributions of the uncertainties to transform the probabilistic safety verification problem to a deterministic safety verification problem and use sum-of-squares based convex methods to verify the obtained deterministic safety constraints over the entire planing time horizon. Provided  approaches,  to  verify  the  safety,  do not  need  uncertainty  samples  and  time  discretization  and are  suitable  for  online  planning  problems  and  long  planing horizons.

\section{Appendix}

\subsection{Proof of Theorem 1}

The probability of not satisfying the safety constraint for a given point $\mathbf{x} \in \mathcal{X}$ at time $t$ is equivalent to the expectation of the indicator function of the uncertain safe set $\mathcal{X}_{s_i}(\omega_i,t)$, i.e., {\small $\hbox{Prob}\left( \mathbf{x} \notin \mathcal{X}_{s_i}(\omega_i,t)    \right)  = \int_{ \{ (\mathbf{x},\mathbf{\omega}_i,t)  \notin \mathcal{X}_{s_i}(\omega_i,t)  \}} pr(\mathbf{\omega}_i)d\mathbf{\omega}_i= \mathbb{E}[\mathbb{I}_{\mathcal{X}_{s_i}}]$},
where the expectation is taken with respect to the 
probability density function of $\mathbf{\omega}_i$ denoted by $pr(\mathbf{\omega}_i)$ \cite{Contour,Risk_Ind}. Also, $\mathbb{I}_{\mathcal{X}_{s_i}}$ is the indicator function of the set $\mathcal{X}_{s_i}(\omega_i,t)$  defined as $\mathbb{I}_{\mathcal{X}_{s_i}} = 1 $ if ${(\mathbf{x},\mathbf{\omega}_i,t)  \notin \mathcal{X}_{s_i}(\omega_i,t)} $, and
0 otherwise.

To find the upper bound of \begin{small}$\hbox{Prob}\left( \mathbf{x} \notin \mathcal{X}_{s_i}(\omega_i,t)    \right) $\end{small}, i.e., polynomial 
\begin{small}
$ P(\mathbf{x},\mathbf{\omega}_i,t)\geq  \mathbb{I}_{\mathcal{X}_{s_i}}$ $\Rightarrow$
$ \mathbb{E}[P(\mathbf{x},\mathbf{\omega}_i,t)] \geq  \mathbb{E}[\mathbb{I}_{\mathcal{X}_{s_i}}]=\hbox{Prob}\left( \mathbf{x} \notin \mathcal{X}_{s_i}(\omega_i,t)    \right)$\end{small}, we need to look for the upper bound approximation of the indicator function $\mathbb{I}_{\mathcal{X}_{s_i}}$.  
In this paper, we will use the upper bound polynomial indicator function and the upper bound probability provided by
\textit{Cantelli's inequality} defined for scalar random variables. For more information see (proof of Theorem 1 in \cite{Contour22}). For other different indicator function-based probability bounds see\cite{Risk_Ind, jasour2019risk}.

Although, the standard Cantelli probability bound addresses scalar random variables and uses first and second moments, the obtained probability bound in this paper addresses multivariate uncertainties, e.g., $\mathbf{x} \in \mathbb{R}^{n_x}$, involving nonconvex and nonlinear sets, e.g., $\mathcal{X}_{s_i}(\omega_i,t)$, and uses higher order moments of the uncertainties. More precisely, It needs the moments of $\omega_i$ up to order $2d$ where $d$ is the order of the polynomial $g_i$.

\subsection{Proof of Theorem 2}

Any state trajectory ${\mathcal{P}}(t) \in \hat{\mathcal{C}}_{r_i}^{\Delta}(t) |_{i=1}^{n_s} \forall t\in[t_0,t_f]$ satisfies the probabilistic safety constraints in \eqref{cc_1}. This implies that polynomials \begin{small}
$\Delta - \frac{P_{1i}(\mathcal{P}(t),t)-P^2_{2i}(\mathcal{P}(t),t)}{P_{1i}(\mathcal{P}(t),t)}, i=1,...,n_s$\end{small} and \begin{small}
$P_{2i}(\mathcal{P}(t),t), i=1,...,n_s$\end{small} should be nonnegative for all $t$ in the planning time interval \begin{small}$\{t: (t-t_0)(t_f-t) \geq 0 \}$\end{small}. To verify the nonnegativity of the polynomials, we use Putinar’s nonnegativity certificate as follows: Polynomial $P(t): \mathbb{R} \rightarrow \mathbb{R} $ is nonnegative on the compact set $\{t: p(t) \geq 0 \}$ if and only if polynomial $P(t)$ can be written as $P(t)=\sigma_0(t) + \sigma_1(t) p(t)$ where $\sigma_0(t)$ and $\sigma_1(t)$ are SOS polynomials \cite{putinar1993positive,SOS2,SOS3}. This results in constraints \eqref{SOS_Cond1} and \eqref{SOS_Cond2}. Note that SOS condition is necessary and sufficient nonnegativity condition for univariate polynomials \cite{SOS2,SOS3}.

\subsection{Proof of Theorem 3}

Any tube ${\mathcal{Q}( \mathcal{P}}(t)) \subset \hat{\mathcal{C}}_{r_i}^{\Delta}(t) |_{i=1}^{n_s} \forall t\in[t_0,t_f]$ satisfies the probabilistic safety constraints in \eqref{cc_t1}. This implies that polynomials \begin{small}
$\Delta - \frac{P_{1i}(\mathcal{P}(t)+\hat{\mathbf{x}}_0,t)-P^2_{2i}(\mathcal{P}(t)+\hat{\mathbf{x}}_0,t)}{P_{1i}(\mathcal{P}(t)+\hat{\mathbf{x}}_0,t)}, i=1,...,n_s$\end{small} and \begin{small}
$P_{2i}(\mathcal{P}(t)+\hat{\mathbf{x}}_0,t), i=1,...,n_s$\end{small} should be nonnegative for all $t$ in the planning time interval \begin{small}$\{t: (t-t_0)(t_f-t) \geq 0 \}$\end{small} and for all $\hat{\mathbf{x}}_0$ in  \begin{small}$\{ \hat{\mathbf{x}}_0: 1-\hat{\mathbf{x}}_0^T{Q}\hat{\mathbf{x}}_0 \geq 0 \}$\end{small}. Note that at each time $t$, we model the tube \begin{small}${\mathcal{Q}( \mathcal{P}}(t))$\end{small} as \begin{small}$\mathcal{P}(t)+\{  \forall \hat{\mathbf{x}}_0: 1-\hat{\mathbf{x}}_0^T{Q}\hat{\mathbf{x}}_0 \geq 0 \}$\end{small}.

To verify the nonnegativity of the polynomials, we use Putinar’s nonnegativity certificate as follows: Polynomial $P(\mathbf{x}): \mathbb{R}^n \rightarrow \mathbb{R} $ is nonnegative on the compact set $\{\mathbf{x} \in \mathbb{R}^n: p_i(\mathbf{x}) \geq 0 \ i=1,...,m\}$ if polynomial $P(t)$ can be written as $P(\mathbf{x})=\sigma_0(\mathbf{x}) + \Sigma_{i=1}^m \sigma_i(\mathbf{x}) p_i(\mathbf{x})$ where $\sigma_i(\mathbf{x}), i=0,...,m$ are SOS polynomials \cite{putinar1993positive,SOS2,SOS3}. This results in constraints \eqref{SOS_t_Cond1} and \eqref{SOS_t_Cond2}.

\bibliographystyle{IEEEtran}
\bibliography{references} 
\end{document}